\documentclass[conference]{IEEEtran}
\IEEEoverridecommandlockouts
\usepackage{cite}
\usepackage{amsmath,amssymb,amsfonts}
\usepackage{algorithmic}
\usepackage{booktabs}
\usepackage{graphicx}
\usepackage[ruled]{algorithm2e}
\usepackage{textcomp}
\usepackage{xcolor}
\usepackage{multirow}
\usepackage{color}
\def\BibTeX{{\rm B\kern-.05em{\sc i\kern-.025em b}\kern-.08em
    T\kern-.1667em\lower.7ex\hbox{E}\kern-.125emX}}

\definecolor{commentcolor}{RGB}{110,154,155}   
\newcommand{\PyComment}[1]{\fontfamily{cmtt}\selectfont\textcolor{commentcolor}{\# #1}}  
\newcommand{\PyCode}[1]{\fontfamily{cmtt}\selectfont\textcolor{black}{#1}} 
    
\begin{document}

\title{SSF3D: Strict Semi-Supervised 3D Object Detection
with Switching Filter
}

\author{\IEEEauthorblockN{1\textsuperscript{st} Songbur Wong}
\IEEEauthorblockA{\textit{School of Aeronautics and Astronautics, SJTU} \\
ShangHai, China \\
songbur\_929@sjtu.edu.cn}
}

\maketitle

\begin{abstract}
SSF3D modified the semi-supervised 3D object detection (SS3DOD) framework, which designed specifically for point cloud data. Leveraging the characteristics of non-coincidence and weak correlation of target objects in point cloud, we adopt a strategy of retaining only the truth-determining pseudo labels and trimming the other fuzzy labels with points, instead of pursuing a balance between the quantity and quality of pseudo labels. Besides, we notice that changing the filter will make the model meet different distributed targets, which is beneficial to break the training bottleneck. Two mechanism are introduced to achieve above ideas: strict threshold and filter switching. The experiments are conducted to analyze the effectiveness of above approaches and their impact on the overall performance of the system. Evaluating on the KITTI dataset, SSF3D exhibits superior performance compared to the current state-of-the-art methods. The code will be released here.
\end{abstract}


\section{Introduction}
\label{sec:intro}
In 3D object detection from LIDAR point clouds, supervised deep learning 
methods currently dominate. 
These approaches require extensive human effort for annotation, training 
and optimization on different scenes, 
limiting their generalization capability.
Semi-supervised learning using teacher-student method \cite{yang2022survey} 
alleviated this problem by
utilizing pseudo labels generated from teacher models. Over the past few years, 
this system 
has been extensively explored and shown great model enhancement 
capabilities\cite{hu2022teacher}. This paper focus on
the semi-supervised learning with teacher-student approach, in particular for 3d object 
detection(3DOD) with 
LIDAR point clouds.

\begin{figure}[t]
  \centering
	\includegraphics[width=1\linewidth]{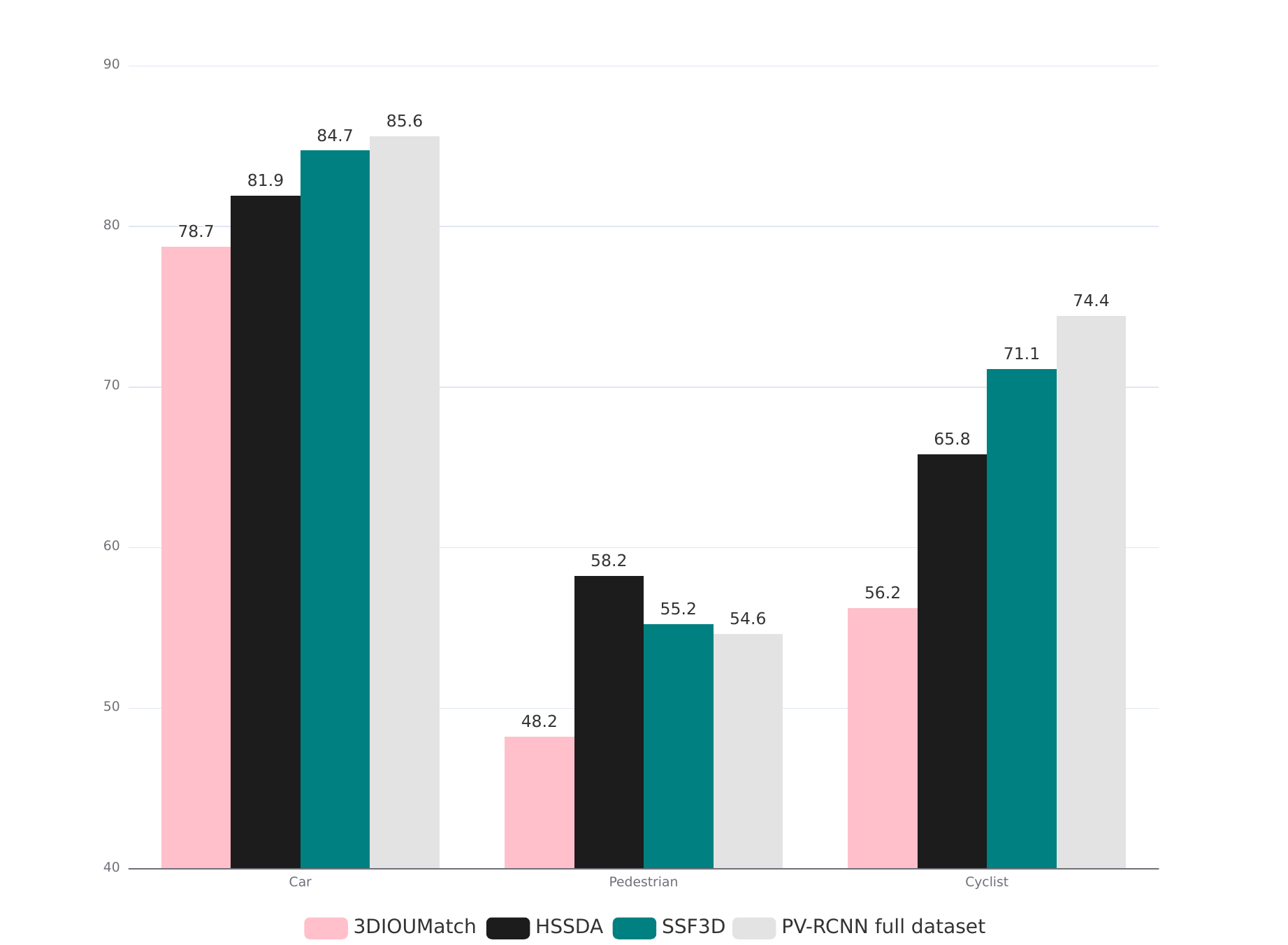} 
	\caption{The proposed SSF3D result compare with state-of-the-art SS3DOD 
  algorithms\cite{wang3DIoUMatchLeveragingIoU2021,liuHierarchicalSupervisionShuffle2023a} using only 2\% labeled KITTI\cite{geigerVisionMeetsRobotics2013} data, and full-dataset supervised result is given in gray bars.}
	\label{plot}
\end{figure}

\begin{figure}[htbp]
	\center{\includegraphics[width=8cm]  {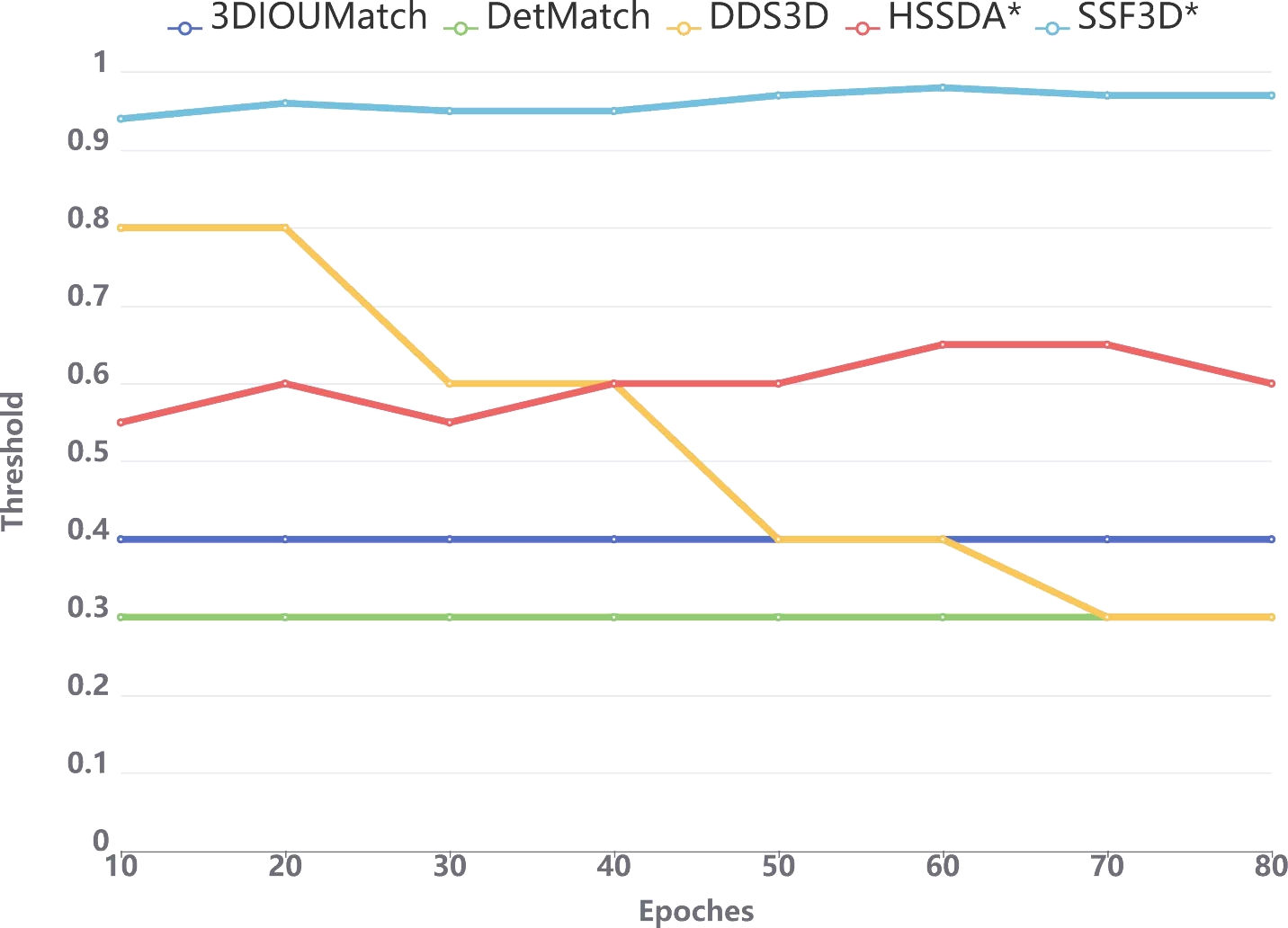}} 
	\caption{\textbf{Threshold picking of different methods.} The confidence thresholds selected by different 
  SS3DOD methods in recent years\cite{wang3DIoUMatchLeveragingIoU2021,parkDetMatchTwoTeachers2022,liDDS3DDensePseudoLabels2023,liuHierarchicalSupervisionShuffle2023a}
   are listed here, noted that those value are for car class, and means the lowest predictive score of the boxes that will 
   be retained as labels. It can be seen that except our method, all other architecture adopt a moderate threshold to make a
    tradeoff between label quantity and quality, meanwhile introduce some incorrect labels.
    The * methods uses a point remove strategy.} 
   \label{threshold}
\end{figure}

\begin{figure}[htbp]
	\center{\includegraphics[width=8cm]  {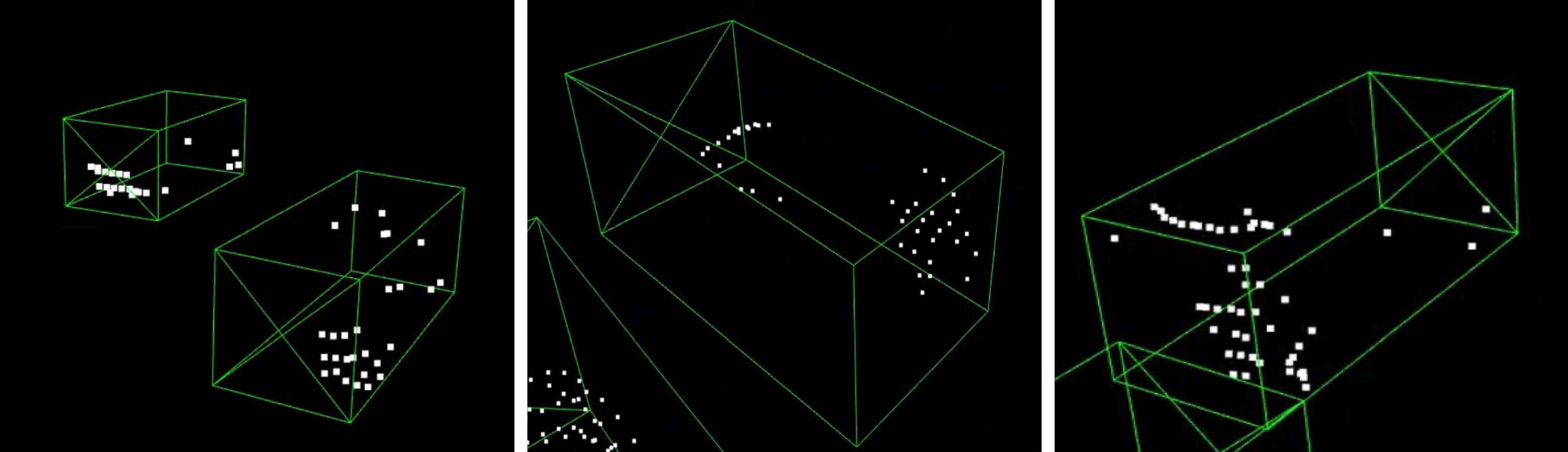}} 
	\caption{\textbf{Ambiguous pseudo labels retained by confidence filters.} In previous works, some
  ambiguous targets that have same points distribution with sundries are remained.
 }
   \label{lt}
\end{figure}

For SS3DOD, the teacher-student framework is used as one of the most
 widely proven semi-supervised learning method. Key idea of this system is to create a 
 performance gap between teacher 
 and student models by applying different augmentations, then use the teacher's outputs 
 as pseudo-labels to supervise the student
  training, which allows exploiting abundant unlabeled data. 
During the training, student model's parameters keep updating, 
 while the teacher's parameters are renew from student's by 
 exponential moving
 average (EMA) strategy commonly.
Recently, the SS3DOD system pioneered by 
3DIOUMatch\cite{wang3DIoUMatchLeveragingIoU2021}
 has led to some 
 works\cite{liDDS3DDensePseudoLabels2023,liuHierarchicalSupervisionShuffle2023a,parkDetMatchTwoTeachers2022}, 
these methods have proposed operators for refining teachers' prediction boxes, while 
in these filtering processes, there still exists a question: 
the retention of erroneous targets generated from teacher network leads 
to interference with training. 
There are two main reasons for this problem, as shown in the Fig. \ref{threshold}
and Fig \ref{lt}.

One is that the current method adopts an imprecise design in the selection 
of filter threshold in order to balance the number and accuracy of retained 
labels. Another is the current filters that rely solely on model confidence 
scores will classify some targets with less information into labels, those 
low information targets are sometimes have same points distribution with other artificial objects,
which tend to be harmful to models training, especially for lightweight ones.

To address the limitation mentioned, we propose a strict teacher-student 
system with switching filters in different stages called SSF3D. 
Following prior works, SSF3D
adopts a point remove operation - eliminating the filtered box along with the 
points inside the box,
due to the existence of this operation, in the tradeoff between the 
precision and the number of labels, it's possible to avoid penalty of the 
model being affected by false negative labels while improving the accuracy 
of the retained labels as much as possible. Specifically, we renew the threshold
selection method to restrict the pseudo-labels to high-confidence boxes, meanwhile
use the Quality Focal Loss \cite{zhouDenseTeacherDense2022} for suppressing the
effects of possible false negative labels. With the help these tricks, the student 
model will only see the points within the high-confidence labels and background 
points during training.


Another attemption to improve the quality of pseudo labels is to switch the
filtering strategy in different stages of training. 
As shown in Fig. \ref{lt},
some targets are regard as labels but it is evident that 
these goals are indistinguishable even with the human eyes.
We believe that this case is caused by the feature distribution similarity, which
means these ambiguous targets have similar latent feature with the true labels.
And in this situation, only use confidence filter for more training is not suitable.

To mitigate this, we propose "Entropy filter" based on
normal consistency, which filter calculates the relationship between box and 
points inside to evaluate the information content 
of a label and filtering against that standard. 
By switch on Entropy filter after several epochs of training with confidence filter,
the model encounter different true positive labels with different features.

With the above strategies, SSF3D achieves a significant improvement on
 the KITTI benchmark \cite{geigerVisionMeetsRobotics2013}, 
comparing to the state-of-the-art HSSDA\cite{liuHierarchicalSupervisionShuffle2023a}. 
Using 1\%, 2\% of labeled data for the car category, SSF3D achieved 
inference mAP increase of 2.5\%, 2.8\% 
respectively under 0.7 IoU at 40 recall.
 For the smaller target class, cyclist, using only 1\% labeled data, SSF3D 
 achieved a 20\% improvement in
inference accuracy over previous approaches. This significant gain demonstrates 
the advantage of SSF3D
 for few-shot detection.

Our main contributions can be summarized as follows:

\begin{itemize}
\item We propose SSF3D, a SS3DOD architecture that is highly compatible with point 
cloud data. Leveraging the sparse and non-overlapping target 
distribution in point cloud frames, SSF3D reduced the artificial disturbs introduced 
in traditional semi-supervised framework.
\item Specifically, we select Gaussian Mixture Modeling and Quality Focal Loss as tools
to conduct a strict training, and introduce a filter-switching strategy to further
enhance label quality. The innovation of strict training and filter-switching strategy 
from the perspective of threshold and filter design reduces the side effects of false 
positive labels.
\item We achieve notably improved performance over the previous state-of-the-art SS3DOD 
method on KITTI dataset with 1\% and 2\% labels, meanwhile two
target level point cloud data augmentation methods are used to conduct experiments and 
perform analysis.
\end{itemize}

\section{Related Work}
\subsection{3D Object Detection with LIDAR Point Clouds}
  
For 3DOD with LIDAR point clouds, major data representations include voxel-based
\cite{s18103337, shi2020points, zhou2022centerformer,langPointPillarsFastEncoders2019,liPillarNeXtRethinkingNetwork2023,
yinCenterbased3DObject2021, zhengCIASSDConfidentIoUAware2020, liu2022bevfusion}
, point-based \cite{qi2019deep, yang20203dssd, xu2021paconv, shi2019pointrcnn}
, and hybrid voxel-point methods \cite{shiPVRCNNPointVoxelFeature2022, shiPVRCNNPointVoxelFeature2021, he2020structure}.
 Architecturally, one-stage \cite{zhengCIASSDConfidentIoUAware2020,
zhou2022centerformer,langPointPillarsFastEncoders2019,liPillarNeXtRethinkingNetwork2023,yang20203dssd}
  and two-stage \cite{shi2019pointrcnn,shiPVRCNNPointVoxelFeature2022,shiPVRCNNPointVoxelFeature2021,deng2021voxel}
 frameworks exist.

 In outdoor scenes, voxel and voxel-point approaches are prevalent since huge raw point sets are computationally 
 expensive. For voxel methods, bird's eye view (BEV) detection is common. PillarNeXt \cite{liPillarNeXtRethinkingNetwork2023}
  achieved high accuracy 
 using only pillar voxel features, PV-RCNN \cite{shiPVRCNNPointVoxelFeature2021}
  exemplifies two-stage techniques - generating 
 proposals in BEV then refining with points. It demonstrated state-of-the-art results at the time by effectively combining 
 both voxel and point features, menawhile remains a strong competitor in 3DOD.

Currently, self-attention models\cite{vaswani2023attention} have become popular in computer vision. For 3DOD, many approaches apply transformers
 to learn intricate features from point clouds \cite{zhou2022centerformer,liu2022bevfusion,fanEmbracingSingleStride2021,
liPillarNeXtRethinkingNetwork2023}.
Transformers excel at capturing complex representations but require abundant labeled datam, 
while their high capacity also enables generating more accurate pseudo-labels, 
which facilitates semi-supervised learning. Thus, transformer-based 3DOD models are also well-suited for semi-supervised paradigms.

Generally speaking, a fine designed semi-supervised learning system can benefit most 3D object detection algorithms, and our SSF3D is designed for this 
purpose. Many pioneering studies have used PV-RCNN\cite{shiPVRCNNPointVoxelFeature2021} as a testbed, so we will also conduct experiments on PV-RCNN 
to evaluate SSF3D. Thanks to OpenPCDet\cite{openpcdet2020}, our semi-supervised methods can be easily applied to various other 3D detection algorithms 
besides PV-RCNN.

\subsection{Semi-supervised Object Detection}
STAC\cite{sohnSimpleSemiSupervisedLearning2020} pioneered teacher-student frameworks for 2D semi-supervised object detection (SSOD), using offline 
training. After that several approaches have driven progress in semi-supervised 2D object detection. Unbiased Teacher\cite{liuUnbiasedTeacherSemiSupervised2021} 
introduced end-to-end training and difference augmentation strategies. Soft Teacher\cite{xuEndtoEndSemiSupervisedObject2021} decoupled the prediction heads 
and proposed box jittering to refine pseudo-labels. Unbiased Teacher v2 \cite{liuUnbiasedTeacherV22022} introduced a Listen2Student mechanism that 
filtered regression labels by uncertainty to enable anchor-free models. Dense Teacher \cite{zhouDenseTeacherDense2022} weighted pseudo-box loss based 
on model confidence scores, and consistent Teacher \cite{wangConsistentTeacherReducingInconsistent2023} prevented label drift via insert a short branch to calibrate 
the predict labels use the classification scores.

The extensive research and experience with 2D data has been leveraged by researchers to be applied to 3D domains, 3DIOUMatch
\cite{wang3DIoUMatchLeveragingIoU2021} set a milestone in SS3DOD, they create an IOU with detect-score guided filter for pseudo
labels.
SE-SSD \cite{zhengSESSDSelfEnsemblingSingleStage2021}  
used a mixup-like 
\cite{yunCutMixRegularizationStrategy2019} technique to combine and mix the 
point clouds of different objects during 
training, meanwhile refined the training loss which brings great benefit. 
Dynamic threshold and dense pseudo labels 
are utilized in DDS3D\cite{liDDS3DDensePseudoLabels2023} to generate more 
correct labels, which enhances the model's performance. However, these 
operations
have some drawbacks during training as they introduce artificially negative 
pseudo labels and can lead to ambiguity in the training process.
DetMatch\cite{parkDetMatchTwoTeachers2022} use a two-teacher system to 
leverage 2D information to boost the SS3DOD, which improved the quality of 
the model but new training complexity is introduced. 
HSSDA\cite{liuHierarchicalSupervisionShuffle2023a} integrates previous works
and proposed a complete offline - online semi-supervised learning architecture.

Above works meanly inherit the 2D semi-supervised learning thought that the
effect of false negative labels is inevitable, so the filtering idea of 
labels is also eclectic to give a better tradeoff.
While in SSF3D, strict filter mechanism replaces the previous eclectic, all
the points possible lead to ambiguous are removed, meanwhile we add a filter
switching strategy to further optimize the pseudo labels.
It's proved that in small
dataset as KITTI, this strategy can bring a significant improvement.
In relation to previous works, we leverage HSSDA's architecture and 
re-designed the filter mechanism, the specific framework will be given 
in Sec. \ref{framework}.

\section{Method}
\subsection{Overall Framework}\label{framework}

SSF3D inherit online-offline alternating training framework
from \cite{liuHierarchicalSupervisionShuffle2023a}, which means
teacher model not only generate pseudo labels during training, but also
save the high scoring boxes with points offline for the convenience of
gt augmentation\cite{s18103337}.
the specific framework is shown in Algorithm \ref{alg1}.

In this framework, GMM3 threshold picking method, Dense Loss and 
Entropy-filter with a switching filter strategy are designs we introduced,
and will be specifically ininterpreted in Sec. \ref{GMM3}, Sec. \ref{DenseLoss} and
Sec. \ref{ENTR}.

\setlength{\algomargin}{0.001in}
\begin{algorithm}[h!]
\SetAlgoLined
\PyComment{\small d: the whole dataset with unlabeled data.} \\
\PyComment{\small d\_l: part of the dataset that is labeled.} \\
\PyComment{\small m\_t, m\_s: teacher and studnet model.} \\
\PyComment{\small GMM3: threshold picking method.} \\
\PyComment{\small Filter: label filter method, include 3 sub-filters for two-stage prediction scores and a custom filter score.} \\
\PyComment{\small dense\_l,reg\_l: classification loss and regression loss for SSF3D.} \\
\PyComment{\small remove\_pts$($d, reslut$)$: functions that remove low score result covered points from data.} \\
\PyComment{\small m: momentum.} \\
\PyComment{\small gt\_aug: random add saved pseudo label into each frame.} \\

\PyCode{\small m\_t, m\_s = pretrain\_model} \PyComment{\small initialize.}\\
\PyCode{\small m\_t.detach$()$} \\
\PyCode{\small Filter = iou\_filter} \PyComment{\small initialize filter.}\\
\PyCode{\small for epoch in total\_epoches:} \\
    \Indp   
    \PyCode{\small if epoch $>$ set\_epoch:} \\
    \Indp
    \PyCode{\small Filter = entropy\_filter} \\
    \PyComment{\small switch filter after several epoches.}\\
    \Indm
    \PyCode{} \\
    \PyCode{\small if epoch $//$ freq == 0:} 
    \PyComment{\small threshold and local label updated every freq epoches.} \\
    \Indp
    \PyCode{\small x\_1 = m\_t$($d\_l$)$, thr = GMM3$($x\_1$)$} \\
    \PyCode{\small x\_2 = m\_t$($d$)$} \\
    \PyCode{\small pseudo\_label = Filter$($x\_2, thr$)$} \\
    \PyCode{\small d.update$($pseudo\_label$)$} \PyComment{\small save the pseudo labels in dataset for future training.} \\
    \Indm 
    \PyCode{} \\
    \PyCode{\small d\_temp = gt\_aug$($d$)$} \\
    \PyCode{\small y\_1 = m\_t$($d\_temp$)$} \\
    \PyCode{\small disposable\_label = Filter$($y\_1, thr$)$} \\
    \PyComment{\small generate single-use label for each epoch} \\
    \PyCode{\small label = concat$($d\_temp$[$label$]$, disposable\_label$)$} \\
    \PyCode{\small d\_temp = remove\_pts$($d\_temp, y\_1$)$} \\
    \PyCode{\small y\_2 = m\_s$($d\_temp$)$} \\
    \PyCode{} \\
    \PyCode{\small loss = dense\_l$($label,y$)$ + reg\_l$($label,y$)$} \\
    \PyCode{\small loss.backward$()$} \\
    \PyComment{\small In the actual calculation, the iteration is carried out in batch} \\
    \PyCode{\small update$($m\_s.params$)$} \\
    \PyCode{\small m\_t.params = m$\times$m\_t.params + $($1-m$)$$\times$m\_s.params} \\
    \Indm
\caption{PyTorch-style pseudo code of SSF3D framework.}
\label{alg1}
\end{algorithm}

\subsection{Strict Mechanism}
Strict Mechanism is a key factor make the SSF3D surpass other 
algorithms. Following previous work, combination of delete and add is 
used in SSF3D. 
Delete means we remove all points from the box whose detective score
does not exceed the filter threshold, in each frame. Add means we save 
the boxes with points that pass the filter, then randomly add them 
into different scene as gt augmentation\cite{s18103337}. Delete operation 
makes student model won't generate loss on ambiguous targets, which help
this framework more robust to the false negative labels. While add
operation ensures the model can reach enough targets to learn.

Therefore, threshold that can determine whether a set of target points 
is deleted or added becomes the key to the design, to remain true positive
pseudo label as correct as possible, the threshold need to be strict but
not extreme. So we design GMM3 
Picker to find this threshold. Simultaneously, we reference Dense 
Loss\cite{zhouDenseTeacherDense2022} to maintain the pseudo label's
quality during generate single-use label for each epoch, these two methods 
work together to ensure a more stringent label filtering than previous
works.

\subsubsection{GMM3 Picker}\label{GMM3}

GMM3 Picker attempt to select suitable thresholds automatically
from data with true labels.
As mentioned in Algorithm \ref{alg1}, all boxes detectived on the labeled 
partial dataset and their corresponding scores will be recorded, each
boxes commonly related to three scores: two scores from two-stage 
detection, and one score from the custom filter. 

Suppose there are $n$ targets in partial dataset with label information, 
and the teacher model contains $m$ detection results for this dataset.
Then calculate the intersection over union (IOU) between $n$ targets
and $m$ detection results, pick the IOU greater than a baseline
\textit{bs\_line=0.5} from detection results as the positive samples. If there 
are $l(l\leqslant m\cap n)$ positive samples,
then there will be a $(l\times 3)$ tensor that record the three scores,
we put each $l$ scores into a 3-level Gaussian Mixture Modeling (GMM) \textit{P}\text{(s)} to search
the correct threshold, which can be described as

\begin{equation}
    P(s)=\sum_{k=1}^3 \omega_k \mathcal{N}\left(s \mid \mu_k, \Sigma_k\right), \quad \sum_{k=1}^3 \omega_k=1 ,
\end{equation}

\noindent where \textit{$\mathcal{N}\left(s \mid \mu_k, \Sigma_k\right)$} 
represent a distribution of each level threshold, and 
\textit{$\omega_k,\mu_k, \Sigma_k $} denote the weight, mean and variance of each level threshold, the mean of each level will be calculated from 
an Expectation-Maximization (EM) algorithm,
after GMM fitting, we generate 3 thresholds from each $l$ scores, and for
a strict filter, we choose the high or middle threshold as the final threshold.
In the end, 3 thresholds will be used to filter the pseudo labels,
results from teacher model which passed all three thresholds will be
set as the final pseudo label.

GMM method can find means of different distributions of the scores, in our case,
there are three means, the lowest one is used to filter the noise, specifically,
during the selection of positive samples, some boxes with low scores will be
selected, and these boxes' scores are likely to be noise. The middle one can be
seen as a moderate threshold, this value is enough high in almost situation but
sometimes fluctuation by input data.
While the highest one mean can generate stable strict threshold, both moderate
and strict threshold are used in SSF3D, commonly, the strict threshold is used
for confident threshold, and the moderate threshold is used for the threshold
of roi score and custom filter.

\subsubsection{Dense Loss}\label{DenseLoss}

Not only focus on selection of pseudo label, we also borrow the  dense 
loss function\cite{zhouDenseTeacherDense2022}
to suppress noise labels. 
After point remove operation,
there still have possibility that some ambiguity points exist in the frame,
those points have similar feature with the true labels, but didn't be
detected by the teacher model, so didn't be removed from the frame.
In this case, once student model recognize these points as labels, it will
harm to the training process. To avoid this situation, the easy solution is
to suppress the effect of low score boxes during training, while dense loss
naturally suitable to complete this mission.

We selected  
dense pseudo loss\cite{zhouDenseTeacherDense2022} as the classification 
loss for two stages of PV-RCNN, which can write as:
\begin{equation}
    \operatorname{\mathcal{L}^{c l s}}=-\left|\hat{y}-y_s\right|^\gamma*
    \left[\hat{y}\log y_s+\left(1-\hat{y}\right)\log\left(1-y_s\right)\right],
\end{equation}

\noindent where \textit{$y_s$} is the predict score of student model, 
and \textit{$\hat{y}$} obey the following rules in our model:

\begin{equation}
        \hat{y}=\left\{\begin{aligned}
        y_t, & \text { if \textit{$y_t>\tau_{p r e d} \ \&\  y_t>\tau_{r o i} \ \&\  y_t>\tau_{thr}$} }, \\
        0, & \text { otherwise },
        \end{aligned}\right.
\end{equation}

\noindent where \textit{$y_t$} is the predict score of teacher 
model, \textit{$\ \tau_{pred},\ \tau_{roi},\ \tau_{thr} $} are thresholds of
 prediction score, roi score and man-designer filter score. 

In situation claired above, \textit{$y_t$} is 0, while the ambiguous
target accidentally found have a low value of \textit{$y_s$}, so the 
nonlinear 
 term\textit{$\ \left|\hat{y}-y_s\right|^\gamma$} can directly suppress
 these point disturb the training.

Combine GMM3 Picker and Dense Loss, strict mechanism in SSF3D let model
only focus on the true positive labels, equivalent to cleaning the data 
at the same time of end-to-end iterative calculation.

\subsection{Entropy Filter}\label{ENTR}

\begin{figure}[t]
  \centering
  \includegraphics[width=0.9\linewidth]{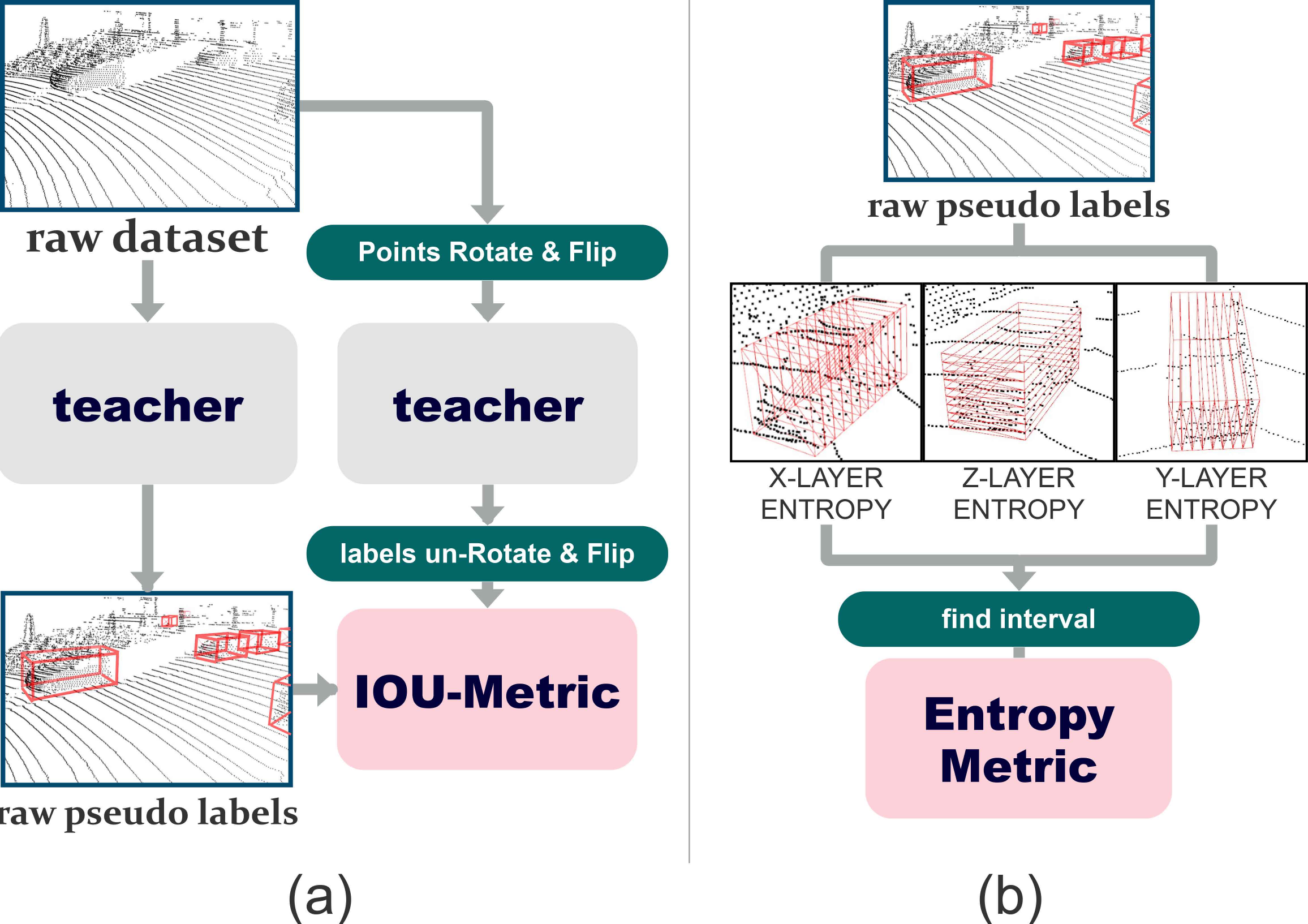}
  \caption{\textbf{Custom filter in 2 stages.} Pipeline of the label 
  filter in SSF3D. 
  in stage-1 training, IOU Filter\cite{liuHierarchicalSupervisionShuffle2023a} 
  as Fig. \ref{Filter} (a) is used.
  In stage-2 we offline count all prediction 
  boxes' 3-dimension-entropy as Fig. \ref{Filter} (b) and use the scatterplot 
  of entropy to find a correct range to filter the 
  generated labels.}
  \label{Filter}
\end{figure}

Each pseudo label have three scores, two scores from two-stage detection,
and one score from the custom filter. The custom filter is also 
one of the key
elements that discriminate a fine-designed SS3DOD from a vanilla one.
We found that with classical filter, which use confidence scores
as metrics can't distinguish some targets, those targets seems have similar
latent feature with the true labels but lack of informational complexity as Fig. \ref{lt},
while these target tend to harm the precision of model, especially lightweight
models.

To solve this problem, we propose an unsupervised operator that combines the 
relationship between points and boxes, further evaluates the 
likelihood of a 3d box to be an annotation by a simple computation, which we 
call 3d information entropy. In order to retain the capacity of filters that 
rely on the model's own judgment while adding the entropy filtering on 
special targets, we adopt a two-step filtering method, which as well 
ensure that the two 
filters will not interfere with each other. 
\begin{figure*}[htbp]
	\center{\includegraphics[width=16cm]  {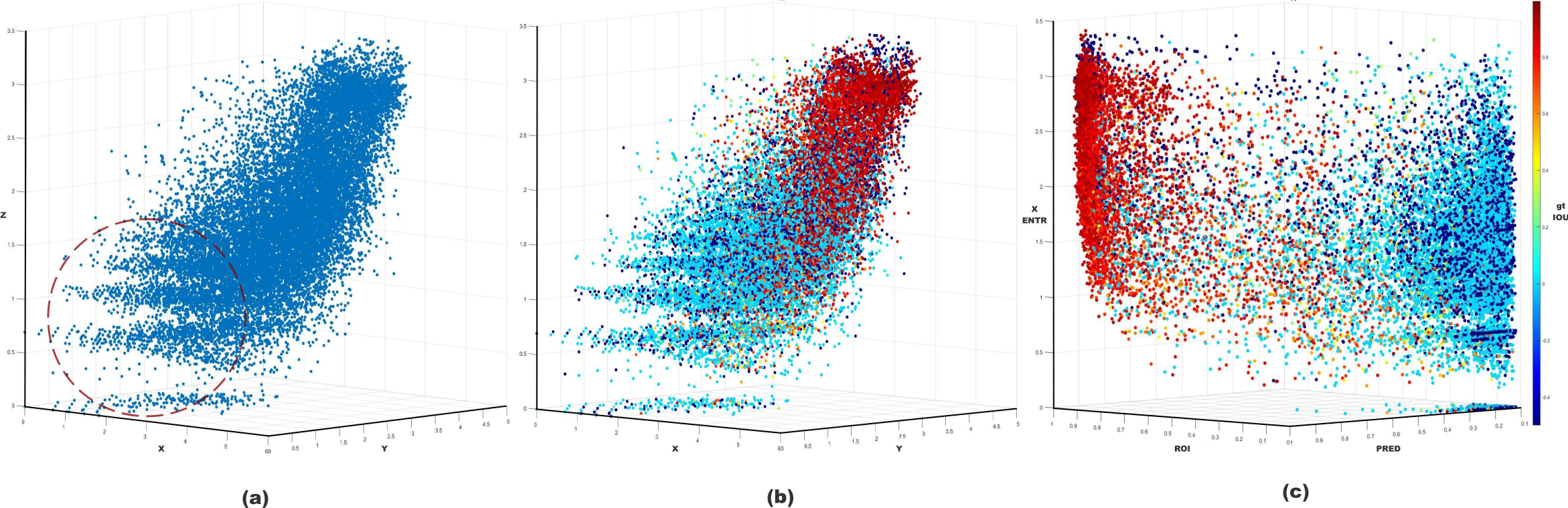}} 
	\caption{\textbf{Scatterplot of all prediction boxes.} Each points in the 
  scatterplot represents a 1-stage-trained teacher model prediction box.
  The darker the red color, the higher the \textit{$IOU_{gt}$}, and the dark blue 
  indicates a classification error.
  (a) The scatterplot of entropy score in target's x,y,z dimension. 
  (b) same scatterplot as Fig.\text{(a)} but with \textit{$IOU_{gt}$} color. 
  (c) The scatterplot of confidence, roi and x-axis entropy score with \textit{$IOU_{gt}$} color.}
   \label{Scatter}
\end{figure*}

As shown in Fig. \ref{Filter}, first step we use the IOU 
filter \cite{liuHierarchicalSupervisionShuffle2023a} to conduct a classical
filtering
, then in second step replacing it by the entropy filter.
Illustrating in Fig. \ref{Filter}\text{(b)}, in entropy filter, a predict box 
is divided into multiple layers according to the spacing \textit{l} for one direction, and the points distributed in the same layer 
are considered co-planar, the 
ratio of points within layer \textit{i} to the total number of points in the box is denoted \textit{$p_i$}, and one dimension entropy can be
 describe as:

\begin{equation}
  \begin{aligned}
  \operatorname{\mathcal{E}_{dim}} = -\sum_{i=1}^L p_{i}\log p_{i} ,
  \end{aligned}
\end{equation}

\noindent where \textit{L} is the total layers of the box in one dimension. 
For different classes, we determine three appropriate thresholds for three dimensions of labels,
 based on their corresponding entropy distribution scatterplot as shown in Fig. \ref{Scatter}\text{(a)}.
The label whose entropy score is lower than the threshold will be filtered out. 
Selection of these thresholds still need human observation, with some trial and error experiments.
To verify the effectiveness of the entropy filter, we plot the scatterplot
of all prediction boxes in Fig. \ref{Scatter}, each scatter in the frame 
represent a detection result, in each scatterplot include all targets that
model detected from 3769 frames of KITTI dataset.

In Fig. \ref{Scatter}\text{(a)} and Fig. \ref{Scatter}\text{(b)}, 
the three axes represent the information
entropy scores in three directions: axis X, axis Y and axis Z.
And the color of each scatter represent the IOU with groundtruth(\textit{$IOU_{gt}$}),
it can be seen that in Fig. \ref{Scatter}\text{(a)}, 
the area circled by the red represents a low score area of entropy,
and it can be verify in Fig. \ref{Scatter}\text{(b)} that this area
indeed contains a large number of false positive labels. For each
class of target, we artificially observe the scatterplot like Fig\text{(a)} and
choose three low score of entropy as the threshold.

We also plot the relationship between the confidence, roi and
entropy score in 
Fig. \ref{Scatter}\text{(c)}, the x,y axis are two scores of prediction
and the z-axis is the entropy score of x direction, it can be seen that
few targets maintain very low entropy when the detection score is high,
and the low entropy targets even in high detection score area, \textit{$IOU_{gt}$}
 is still not high enough to be a label.


In traditional 3D encoders\cite{4650967,5152473,10.1007/978-3-642-15558-1_26}, 
the proximity point is used as a unit to compute the normal 
and encode the point by the relationship between the 3d point normals, 
whereas our layer-wise computation also equivalent to 
measuring the difference between the normals of the points inside the box and
  the normals of the box boundaries, further estimating the accuracy of the 
  regression box. Compare to traditional encoders, information entropy
  computation is more specialize in low-level, introduces more prior 
  information and aims to filter out a 
  specific portion of the low-information box, 
  such as some of the walls that have been misdiagnosed as vehicles and so on.

\section{Experiments}

\subsection{Experiments Setups}


\noindent\textbf{Datasets and Evaluation Metrics.}
Following the methods raised recent years
\cite{liuHierarchicalSupervisionShuffle2023a,liDDS3DDensePseudoLabels2023,wang3DIoUMatchLeveragingIoU2021,parkDetMatchTwoTeachers2022}, 
we use KITTI 3D bench mark\cite{geigerVisionMeetsRobotics2013} as dataset and PV-RCNN as test bed, there are 7481 outdoor scenes for training and 7518 for testing, 
the training 7481 frames have annotations and is generally used in LiDAR algorithm develop, follow lots of prior researches, we use the divided 
\textit{train} split
of 3,712 samples and \textit{val} split of 3,769 samples as a common practice. Using the released 
3DIoUMatch\cite{wang3DIoUMatchLeveragingIoU2021} splits for 
1\% and 2\% of labeled scenes over KITTI train split, We put the proposed theoretical approach into practice. In KITTI,
 the detection target are divided in hard, moderate and easy 3 levels by the ease of testing, in order to make 
a comprehensive comparison with the methods that have emerged in recent years, we use cars, pedestrians and cyclists' mean Average Precision (mAP)
with 40 recall positions obtained at 0.7, 0.5, and 0.5 3D IoU threshold respectively as the comparative indicators.

\noindent\textbf{Implementation Details.}
In SSF3D, all the model basic setting are same 
as PV-RCNN\cite{shiPVRCNNPointVoxelFeature2021} from OpenPCDet\cite{openpcdet2020}, meanwhile weak augmentation 
and the EMA momentum factors are also same as prior works\cite{wang3DIoUMatchLeveragingIoU2021,liuHierarchicalSupervisionShuffle2023a}. 
Before the first epoch and each 5 epochs, teacher model will calculate dataset offline to generate pseudo labels as gt database to implement gt sample 
data-augmentation for further training.

For GMM3 Picker, we set a minimal threshold value for cyclist and pedestrian categories because both of them have a low initial accuracy
and lack of data in KITTI dataset, this minimal threshold
is set from the final value obtained by the GMM3 Picker method after one stage training without minimal threshold. Meanwhile
we use moderate GMM3 thresholds for cyclist and pedestrian categories' filters to ensure the model can reach more pseudo labels.
For Dense Loss, we set \textit{$\gamma$} as \text{2}.
In second stage of SSF3D, we start our training from a well performed checkpoint from stage-1 80 epochs training,
entropy filter's layer length \textit{l} is set as 0.06m, 
and the entropy threshold in 3 dimensions for car category is set as \text{0.9} after observation from Fig. \ref{Scatter} \text{(a)}, while for pedestrian and
cyclist we didn't change their filter due to random normal variation of small target. One NVIDIA GTX3090 is used for all the training.

\subsection{Main Results}

\begin{table}[htbp]
  \centering
    \setlength{\tabcolsep}{1.5mm}{
    \scalebox{0.75}{
    \begin{tabular}{c|r|r|r|r|r|r|r|r}
    \toprule
    \multirow{2}[1]{*}{Model} & \multicolumn{1}{c}{} & \multicolumn{2}{c}{1\%} &       & \multicolumn{1}{c}{} & \multicolumn{2}{c}{2\%} &       \\
          & \multicolumn{1}{c}{Car} & \multicolumn{1}{c}{Ped.} & \multicolumn{1}{c}{Cyc.} & \multicolumn{1}{c|}{Avg.} & \multicolumn{1}{c}{Car} & \multicolumn{1}{c}{Ped.} & \multicolumn{1}{c}{Cyc.} & \multicolumn{1}{c}{Avg.} \\
          \hline
    PV-RCNN\cite{shiPVRCNNPointVoxelFeature2021} & \multicolumn{1}{c|}{73.5} & \multicolumn{1}{c|}{28.7} & \multicolumn{1}{c|}{28.4} & \multicolumn{1}{c|}{43.5} & \multicolumn{1}{c|}{76.6} & \multicolumn{1}{c|}{40.8} & \multicolumn{1}{c|}{45.5} & \multicolumn{1}{c}{54.3} \\
    3DIOUMatch\cite{wang3DIoUMatchLeveragingIoU2021} & \multicolumn{1}{c|}{76.0} & \multicolumn{1}{c|}{31.7} & \multicolumn{1}{c|}{36.4} & \multicolumn{1}{c|}{48.0} & \multicolumn{1}{c|}{78.7} & \multicolumn{1}{c|}{48.2} & \multicolumn{1}{c|}{56.2} & \multicolumn{1}{c}{61.0} \\
    3DDSD\cite{yang20203dssd} & \multicolumn{1}{c|}{76.0} & \multicolumn{1}{c|}{34.8} & \multicolumn{1}{c|}{38.5} & \multicolumn{1}{c|}{49.7} & \multicolumn{1}{c|}{78.9} & \multicolumn{1}{c|}{49.4} & \multicolumn{1}{c|}{53.9} & \multicolumn{1}{c}{60.7} \\
    DetMatch\cite{parkDetMatchTwoTeachers2022} & \multicolumn{1}{c|}{77.5} & \multicolumn{1}{c|}{\textbf{57.3}} & \multicolumn{1}{c|}{42.3} & \multicolumn{1}{c|}{59.0} & \multicolumn{1}{c|}{78.2} & \multicolumn{1}{c|}{54.1} & \multicolumn{1}{c|}{64.7} & \multicolumn{1}{c}{65.6} \\
    HSSDA\cite{liuHierarchicalSupervisionShuffle2023a} & \multicolumn{1}{c|}{80.9} & \multicolumn{1}{c|}{51.9} & \multicolumn{1}{c|}{45.7} & \multicolumn{1}{c|}{59.5} & \multicolumn{1}{c|}{81.9} & \multicolumn{1}{c|}{58.2} & \multicolumn{1}{c|}{65.8} & \multicolumn{1}{c}{68.6} \\
    Our SSF3D & \multicolumn{1}{c|}{\textbf{83.4}} & \multicolumn{1}{c|}{54.3} & \multicolumn{1}{c|}{\textbf{65.6}} & \multicolumn{1}{c|}{\textbf{67.7}} & \multicolumn{1}{c|}{\textbf{84.7}} & \multicolumn{1}{c|}{\textbf{58.6}} & \multicolumn{1}{c|}{\textbf{71.1}} & \multicolumn{1}{c}{\textbf{71.4}} \\
    \bottomrule
    \end{tabular}
    }
    }
    \caption{Comparisons were made between our experimental findings on the KITTI dataset and the most recent state-of-the-art methods. 
    These results were documented across 40 recall positions, employing IoU thresholds of 0.7, 0.5, and 0.5 for 'Car', 'Pedestrian', and 'Cyclist', respectively.}
  \label{main}
\end{table}

\begin{table}[htbp]
  \centering
  \setlength{\tabcolsep}{0.8mm}{
    \scalebox{0.7}{
    \begin{tabular}{c|c|c|c|c|c|c|c|c|c}
    \toprule
    \multirow{2}[2]{*}{Model} & \multicolumn{3}{c|}{Car} & \multicolumn{3}{c|}{Pedestian} & \multicolumn{3}{c}{Cyclist} \\
          & \multicolumn{1}{c}{Easy} & \multicolumn{1}{c}{Mod} & Hard  & \multicolumn{1}{c}{Easy} & \multicolumn{1}{c}{Mod} & Hard  & \multicolumn{1}{c}{Easy} & \multicolumn{1}{c}{Mod} & Hard \\
    \midrule
    Voxel-RCNN* & 88.08 & 76.15 & 71.58 & 47.40 & 42.18 & 37.35 & 62.55 & 44.74 & 41.30 \\
    HSSDA* & 90.26 & 76.76 & 71.57 & 57.74 & 49.65 & 43.49 & \textbf{73.24} & 51.82 & 48.71 \\
    SDF3D & \textbf{90.54} & \textbf{79.70} & \textbf{74.96} & \textbf{60.84} & \textbf{53.77} & \textbf{47.52} & 71.18 & \textbf{52.92} & \textbf{49.35} \\
    \bottomrule
    \end{tabular}%
    }
  }
  \caption{Results comparison on the KITTI dataset using the Voxel-RCNN detector as test bed, used 100 epochs training, 
  used 2\%label dataset, metrics are same as Table \ref{main}.
  * means reproduced by us.}
  \label{voxel}%
\end{table}%

We evaluated several recent methods on the KITTI dataset and their results are 
summarized in Tab. \ref{main}. 
It is evident that SSF3D consistently achieves high accuracy across various 
initial labeled dataset splits. 
In concrete terms, our approach significantly enhanced model performance, 
the average mAP have a remarkable 
increase of nearly 8\% and 5\% for annotation splits of 1\% and 2\%, respectively. 
The results of full-labeled PV-RCNN from the PCDet provided checkpoint 
are also included in the Tab. \ref{ablation}. We consider that for pedestrian and cyclist categories,
the reason why the performance of SSF3D and several previous works better than full-dataset result are similar,
is that these semi-supervised methods filter out some ambiguous labels in the initial.

At the same time, we used Voxel-RCNN\cite{deng2021voxel} 
as a test and placed the results in Table \ref{voxel} to demonstrate the 
portability of our method for any other LIDAR point cloud based algorithm.
We reproduced all the experiments with a mini-batch as 2 for adapting our GPU memory, and the results show that SSF3D
still have a improvement over the state-of-the-art methods.

\subsection{Ablation Study}

All the experiments in this Section are tested on the 2\% KITTI dataset, same metric as Tab. \ref{main}, and the total ablation 
test form is shown in Tab. \ref{ablation}.
\begin{table*}[t]
  \centering
  \setlength{\tabcolsep}{3.7mm}{
    \scalebox{0.8}{
    \begin{tabular}{c|c|c|c|c|c|c|c|c|c|c|c|c}
    \toprule
    \multirow{2}[2]{*}{} & \multirow{2}[2]{*}{GMM} & \multirow{2}[2]{*}{DSL} & \multirow{2}[2]{*}{ENF} & \multicolumn{3}{c|}{Car} & \multicolumn{3}{c|}{Ped.} & \multicolumn{3}{c}{Cyc.} \\
          &       &       &       & \multicolumn{1}{c}{Easy} & \multicolumn{1}{c}{Mod} & Hard  & \multicolumn{1}{c}{Easy} & \multicolumn{1}{c}{Mod} & Hard  & \multicolumn{1}{c}{Easy} & \multicolumn{1}{c}{Mod} & Hard \\
    \midrule
    Exp.1 & -     & -     & -     & 91.36 & 79.53 & 74.63 & 57.86 & 51.84 & 46.72 & 79.37 & 61.04 & 57.24 \\
    Exp.2 & $\surd$     & -     & -     & 92.29 & 81.77 & 77.38 & 65.89 & 58.21 & 52.01 & 90.43 & 65.93 & 62.05 \\
    Exp.3 & $\surd$     & $\surd$     & -     & 93.09 & 82.18 & 77.42 & \textbf{66.27} & \textbf{58.35} & \textbf{52.24} & \textbf{91.58} & \textbf{66.31} & \textbf{62.87} \\
    Exp.4 & $\surd$     & $\surd$     & $\surd$     & \textbf{93.12} & \textbf{82.48} & \textbf{79.38} & 64.38 & 56.02 & 49.35 & 90.66 & 66.24 & 62.43 \\
    {\color{teal} 100\% labed} & -     & -     & -     & {\color{teal} 91.38} & {\color{teal} 82.86} & {\color{teal} 81.92} & {\color{teal} 62.43} & {\color{teal} 53.17} & {\color{teal} 48.44} & {\color{teal} 87.89} & {\color{teal} 69.71} & {\color{teal} 65.53} \\
    \bottomrule
    \end{tabular}%
    }
  }
  \caption{"GMM" means GMM3 Picker module, "DSL" means Dense Loss and "ENF" 
  means filter change with entropy filter. According to the results of this table, we conduct switch of entropy filter 
  only on car category.}
  \label{ablation}%
  \end{table*}%

\noindent\textbf{Effect of GMM3 Picker.}
\begin{table}[htbp]
  \centering
  \setlength{\tabcolsep}{3.7mm}{
    \scalebox{0.9}{
    \begin{tabular}{c|c|c|c|c}
    \toprule
    threshold  & con  & roi   & cus & Car \\
    \midrule
    fixed high & 0.99  & 0.97  & 0.9   & 79.6 \\
    fixed low & 0.5   & 0.5   & 0.5   & 76.7 \\
    HSSDA\cite{liuHierarchicalSupervisionShuffle2023a} & 0.41  & 0.42  & 0.91  & 81.9 \\
    GMM3(our)  & 0.96  & 0.91  & 0.88  & \textbf{83.3} \\
    \bottomrule
    \end{tabular}%
    }
  }
\caption{\textbf{The effect of different threshold selection on the model accuracy.}
For car category, we list the first epoch's three thresholds and the final average mAP score of the model,
these threshold represent the lowest score of a label that can be learned by the model; "con" means confidence score, 
"roi" means roi score and "cus" means custom filter score.}
\label{gmm_com}
\end{table}
The GMM3 Picker is used during both stage, while the threshold 
changing curve can be seen in Fig. \ref{threshold}.
As shown in Tab. \ref{ablation} Exp. 2, after a suitable threshold have been choice, 
detection mAP improved by 2\% in the cars category and 7\% 
in the cycles category. 

We also set up some comparative experiments to prove the influence of 
different threshold selection on the model accuracy. As shown in the Tab. \ref{gmm_com},
compare to HSSDA, who also use a point remove strategy and automatically threshold picking
, we generate more strict but not extreme threshold and get a better result.
At the same time, Tab. \ref{gmm_com} proves that too high and too low thresholds are meaningless 
even for the method with a point remove strategy.

\noindent\textbf{Effect of Dense Loss.}
Dense loss is also used in all stage in SSF3D, from Tab. \ref{ablation} Exp. 3, it can be seen that
this function played an auxiliary role in strict semi-supervision mechanism. It does not bring as 
great an effect as the GMM3 Picker, but it also has a positive effect.

 \noindent\textbf{Effect of Entropy Filter.}\label{ablen}
\begin{table}[htbp]
  \centering
  \setlength{\tabcolsep}{1.7mm}{
  \scalebox{0.9}{
    \begin{tabular}{c|c|c|c|c|c}
    \toprule
    entr\_x & entr\_y  & entr\_z  & Car\_easy & Car\_mid & Car\_hard \\
    \midrule
    1.0     & 0     & 0     & 93.02 & 82.37 & 78.72 \\
    0     & 1.0     & 0     & 1     & 1     & 1 \\
    0     & 0     & 0.9   & 92.59 & 82.39 & 79.29 \\
    1.0   & 1.0   & 0.9   & 93.12 & 82.48 & 79.38 \\
    \bottomrule
    \end{tabular}%
  }
  }
  \caption{\textbf{Comparison of entropy filters applied on different axes.}
  We apply entropy filter on different axes and the result is shown in the table.}
  \label{entr_abl}%
\end{table}%
Entropy Filter is used in the second stage of SSF3D,
as Tab. \ref{ablation} Exp. 4, after 40 epochs Stage-2 training, the car 
category has a boost with 2\% increase in the hard level object. 
While pedestrian and cyclist categories both have some minor fluctuations, 
we believe this 
results from the fact that the point normals of small targets vary 
dramatically, making their information entropy ineffective in identifying the 
validity of some of the prediction frames. Meanwhile the initial labels of 
pedestrian and cyclist categories are very few, thus there are few fuzzy semantic labels
need to filter by entropy filter.
Therefore in the main structure of SSF3D, we did not implement filter changes 
for cyclist and pedestrian, only on the car category.

Further, we conduct some experiments for the selection of the entropy thresholds,
and the results are shown in Tab. \ref{entr_abl}, according to the results, the entropy filter
in z-axis has the most significant impact on the model's performance, we consider that this is because 
most of ambiguous labels have a low entropy score in z-axis. For example, the point cloud z-axis distribution 
of a distant target will naturally be more sparse than a near target, and it will have a higher 
probability of return noise with occlusion.

Information entropy as a filtering lacks of filtering 
accuracy compared to IOU filters designed based on confidence scores,
but it is sufficient to filter out ambiguous targets that are difficult to
distinguish from the model's own judgment, what's more, the latent distribution of
pseudo labels is also slightly changed by filter switching.

\section{Conclusion}
In this paper, we have made a series of explorations in the SS3DOD training framework and proposed
 a two-stage training architecture: SSF3D.
In both stages GMM3 Picker and Dense Loss
 has been introduced to conduct a semi-supervised training with fewer interference from false-negative/positive labels.
  In Stage 2 a new filter has been introduced to obtain different distribution 
objects as labels, meanwhile filter out ambiguous labels which have lower information entropy. 

The experiments confirm that the architecture of deleting and adding is effective,
deleting noise points in low score labels eliminated disturb from false-negative labels and
adding high score labels from the pseudo labels database keeps the complexity of scenes.
Based on this architecture, conduct a strict filter of pseudo labels significantly improved the model's performance,
and GMM3 Picker, Dense Loss, Entropy filter proposed in this paper can helpfully achieve this goal.

{\small
\bibliographystyle{ieeetr}      
\bibliography{ssf3d}    
}

\end{document}